\documentclass[letterpaper, 10 pt, conference]{styles/ieeeconf}
\IEEEoverridecommandlockouts 
\usepackage{multicol}
\usepackage[bookmarks=true]{hyperref}

\usepackage{url}
\usepackage{color}
\usepackage{wrapfig}
\usepackage{subcaption}
\newif\iffigs
\figstrue 

\usepackage[fleqn]{amsmath}
\usepackage{float}
\usepackage{setspace}
\usepackage{graphicx}
\usepackage{mathrsfs}
\usepackage{amssymb}
\usepackage{nicefrac}
\usepackage{algorithm}
\usepackage{caption}
\usepackage[noend]{algpseudocode}

\usepackage{tikz}
\usetikzlibrary{decorations.pathreplacing,calc}

\usepackage{flushend}

\makeatletter
\newcommand\fs@spaceruled{\def\@fs@cfont{\bfseries}\let\@fs@capt\floatc@ruled
  \def\@fs@pre{\vspace{0.4\baselineskip}\hrule height.8pt depth0pt \kern2pt}%
  \def\@fs@post{\vspace{-0.4\baselineskip}\kern2pt\hrule\relax\vspace{-12pt}}%
  \def\@fs@mid{\kern2pt\hrule\kern2pt}%
  \let\@fs@iftopcapt\iftrue}
\makeatother

\usepackage{lipsum}
\newcommand\myeq{\mkern1.5mu{=}\mkern1.5mu}

\usepackage{xfrac}
\newif\ifanon

\title{\LARGE \bf GATO: GPU-Accelerated and Batched Trajectory Optimization\\for Scalable Edge Model Predictive Control}

\ifanon
\else
\author{Alexander Du$^{1}$, Emre Adabag$^{1,2}$, Gabriel Bravo-Palacios$^{3,4}$, Brian Plancher$^{3,4}$
\thanks{This material is based upon work supported by the National Science Foundation (under Awards 2246022, 2411369). Any opinions, findings, conclusions, or recommendations expressed in this material are those of the authors and do not necessarily reflect those of the funding organizations.}
\thanks{$^{1}$ School of Engineering and Applied Science, Columbia University.}
\thanks{$^{2}$ University of Michigan}%
\thanks{$^{3,4}$ Barnard College, Columbia University and Dartmouth College}%
\thanks{Correspondence to: {\tt\footnotesize plancher@dartmouth.edu}}
}
\fi

\begin{document}
\maketitle
\thispagestyle{empty}
\pagestyle{empty}


\begin{abstract}
    While Model Predictive Control (MPC) delivers strong performance across robotics applications, solving the underlying (batches of) nonlinear trajectory optimization (TO) problems online remains computationally demanding. Existing GPU-accelerated approaches either parallelize single solves, handle large batches at sub-real-time rates, or sacrifice model generality for speed. This leaves a large gap in solver performance for many state-of-the-art MPC applications that require real-time batches of tens to low-hundreds of solves. As such, we present GATO, an open source, GPU-accelerated, batched TO solver co-designed across algorithm, software, and computational hardware to deliver real-time throughput for these moderate batch size regimes. Our approach leverages a combination of block-, warp-, and thread-level parallelism within and across solves for ultra-high performance. We demonstrate the effectiveness of our approach through a combination of: simulated benchmarks showing speedups of $18-21\times$ over CPU baselines and $1.4-16\times$ over GPU baselines as batch size increases; case studies highlighting improved disturbance rejection and convergence behavior; and finally a validation on hardware using an industrial manipulator. We open source GATO to support reproducibility and adoption.
\end{abstract}

\section{Introduction} \label{sec:intro}
Model Predictive Control (MPC) is a feedback control strategy which has seen great success in a wide variety of robotic applications~\cite{hogan2018reactive,sleiman2021unified,tranzatto2022cerberus,wensing2023optimization,Kuindersma23Talk}. Most implementations of (nonlinear) MPC leverage trajectory optimization (TO)~\cite{betts2010practical} to solve the underlying optimal control problems. Historically, these TO problems are solved through 1st- or 2nd-order optimization methods. Unfortunately, such problems are computationally expensive and only deliver locally optimal solutions. As such, several recent efforts have leveraged careful approximations and simplifications of the underlying optimal control problem~\cite{kuindersma2016optimization,li2024cafe,nguyen2024tinympc}, as well as hardware acceleration, most commonly on GPUs, to help overcome these computational limitations. These GPU-accelerated efforts include both the development of 0th order methods that construct sample-based approximate gradients~\cite{williams2017model,vlahov2024mppi,xue2025full,alvarez2025real,enrico2025comparison} as well as hybrid, 1st-, and 2nd-order methods targeting both the overall solvers~\cite{plancher2018performance,pan2019gpu,lee22parallelILQR,cole2023exploiting,sundaralingam2023curobo,shin2024accelerating,adabag2024mpcgpu,lee2024gpu,chari2024fast,jeon2024cusadi,bishop2024relu,tracy2024differentiability,amatucci2025primal}, as well as the underlying numerical linear algebra and physics kernels~\cite{naumov2011incomplete,schubiger2020gpu,plancher2021accelerating,plancher2022grid,pacaud2023accelerating}. Importantly, this collection of works demonstrates robust, real-time, real-world usability though numerous deployments onto various modalities of physical robot hardware.

At the same time, there have been a number of recent applications in which batches of tens to low-hundreds of trajectory optimization solves can be leveraged for state-of-the-art MPC performance~\cite{lee2024gpu,amatucci2025primal,hamer2018fast,guhathakurta2022fast,rastgar2023gpu,tsikelis2024gait,lew2025risk}. And while many of these results are demonstrated through GPU parallelism, in general, whether through 0th-, 1st-, 2nd-order, or hybrid approaches, existing GPU-accelerated solvers are designed to either parallelize a single solve across a GPU, implement large-scale (e.g., $>$1000) parallel batches of solves, or are special cased for a very limited setup. As such, to the best of the authors’ knowledge, for batches of tens to low-hundreds of solves, prior solvers trade off latency, throughput, and generality: some hit kHz rates but only for a few parallel solves; others process large batches but miss real-time targets; still others attain speed by restricting the problem specification (e.g., point-mass models, a single linearization). This fundamentally limits their deployed use, despite their demonstrated real-world promise.

\iffigs
\begin{figure}[!t]
    \centering   
    \includegraphics[width=\ifanon0.9\else0.6\fi\columnwidth]{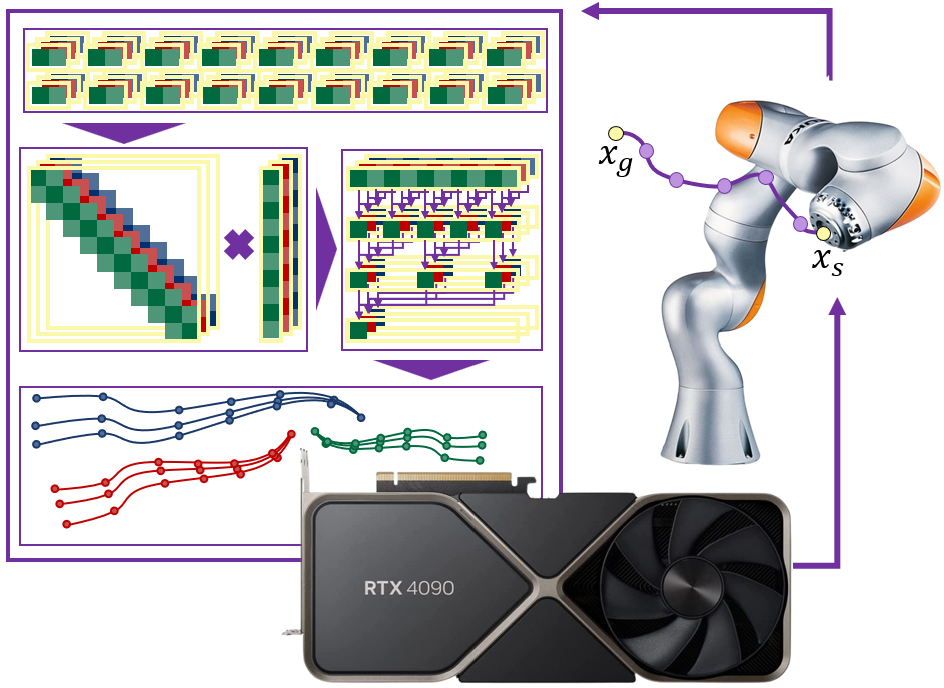}
    \caption{The GATO solver parallelizes across batches of trajectory optimization solves on the GPU through algorithm-software-hardware co-design. This approach enables real-time performance for batch sizes of tens to low-hundreds of solves with tens to low-hundreds of knot points per solve.}
    \label{fig:overview}
    \vspace{-15pt}
\end{figure}
\fi

To overcome these challenges, we developed GATO (Figure~\ref{fig:overview}), a GPU-accelerated, batched trajectory optimization solver designed to enable real-time batched solves of tens to low-hundreds of trajectory optimization problems. Our work is inspired by the MPCGPU solver~\cite{adabag2024mpcgpu}, which demonstrated that GPU-acceleration through careful algorithm-hardware-software co-design can enable long-horizon, real-time performance. While MPCGPU is limited to a single solve per GPU, we designed GATO to solve tens to low-hundreds of problems simultaneously. This is done via block-, warp-, and thread-level parallelism both across and within underlying computations for efficient problem matrix formation, linear system solves, and line search iterate computations.

We demonstrate the power of this GPU-first framework through a series of simulation benchmarks, case studies, and a hardware demonstration on an industrial manipulator. We find that GATO provides speedups of up to $18-21\times$ over CPU baselines and $1.4-16\times$ over GPU baselines as batch size increases. Our case studies highlight how such batched solves can improve disturbance rejection and convergence behavior of TO and MPC and can run in real-time on robot hardware. We release our software open source\ifanon for the benefit of the robotics community\fi~at: 
\ifanon
{\footnotesize \href{https://anonymous.4open.science/r/GATO-ICRA-26}{\color{blue} \texttt{https://anonymous.4open.science/r/GATO-ICRA-26}}}.
\else
{\small \href{https://github.com/a2r-lab/GATO}{\color{blue} \texttt{https://github.com/a2r-lab/GATO}}}.
\fi

\section{Background} \label{sec:background}
\subsection{Direct Trajectory Optimization}
\label{sec:background:trajopt}
Trajectory optimization~\cite{betts2010practical} solves an (often) nonlinear optimization problem to compute a robot's path through an environment as a series of states $X\myeq \{x_0,\dotsi,x_N\}$ and controls $U\myeq  \{u_0,\dotsi,u_{N-1}\}$ for $x$ $\in \mathbb{R}^n$ and $u$ $\in \mathbb{R}^m$. These problems model the robot as a discrete-time dynamical system, $x_{k+1} = f(x_k,u_k,h)$, with initial condition $x_0 = x_s$ and timestep $h$, and minimize an additive cost function, $J(X,U)$.
Recent work has shown that direct methods, which explicitly represent the states, controls, dynamics, and any additional constraints, lead to moderately large nonlinear programs with structured sparsity patterns~\cite{nocedal1999numerical}. These approaches can be greatly accelerated on the GPU, especially as the size of the problem increases~\cite{adabag2024mpcgpu}. Direct methods follow a three-step process which is repeated until convergence~\cite{nocedal1999numerical,wachter2006implementation,gill2005snopt}:

\emph{Step 1:} Form the following quadratic program via a second-order Taylor expansion of the problem along a nominal trajectory, where $Q$, $R$ and $q$, $r$ are the Hessians and gradients of $J$, and $A$ and $B$ are the gradients of $f$, with respect to $x$ and $u$, and $e_k = f(x_k,u_k,h) - x_{k+1}$:
    \begin{equation} \label{eq:trajoptQP}
    \begin{split}
        \min_{\substack{\delta X, \delta U}} 
            &\;\; \tfrac{1}{2}\delta x_N^T Q_N \delta x_N + q_N^T \delta x_N + \\ 
            \sum_{k=0}^{N-1} &\tfrac{1}{2}\delta x_k^T Q \delta x_k + q^T \delta x_k + \tfrac{1}{2}\delta u_k^T R \delta u_k + r^T \delta u_k,\\
        \text{s.t.\quad} &\delta x_0 = x_s - x_0, \\
        \delta x_{k+1}& - A_k \delta x_k - B_k \delta u_k = e_k, \; \forall k = 0, \ldots, N{-}1.
    \end{split}
    \end{equation}
\emph{Step 2:} Compute $\delta X^*, \delta U^*$ by solving the KKT system:
{\renewcommand{\arraystretch}{0.8}
\begin{equation} \label{eq:KKT}
\begin{split}
    &\begin{bmatrix} G & C^T \\ C & 0 \end{bmatrix} \begin{bmatrix} -\delta Z \\ \lambda \end{bmatrix} = \begin{bmatrix} g \\ c \end{bmatrix},\\
\end{split}
\end{equation}}
where $\delta z_k = \begin{bmatrix} \delta x_k & \delta u_k\end{bmatrix}^T$, $\delta z_N = \delta x_N$,
\begin{equation*} \label{eq:KKTSimplifications}
{\renewcommand{\arraystretch}{0.8}
\scalebox{0.99}{$
\begin{split}
    G &= \begin{bmatrix}
        Q_0 & & & \\
        & R_0 & & \\
        & & \ddots & \\
        & & & Q_N \\
        \end{bmatrix}_{\textstyle \raisebox{2pt}{,}} \\
    g &= \begin{bmatrix} q_0 & r_0 & q_1 & r_1 & \dots & q_N \end{bmatrix}^T, \\
    C &= \begin{bmatrix}
           I &      &        &          &          & \\
        -A_0 & -B_0 & I      &          &          & \\
             &      & \ddots & -A_{N-1} & -B_{N-1} & I \\
        \end{bmatrix}_{\textstyle \raisebox{2pt}{,}} \\
    c &= \begin{bmatrix} x_s -  x_0 & e_0 & e_1 & \dots & e_{N-1} \end{bmatrix}^T_{\textstyle \raisebox{2pt}{.}}
\end{split}
$}}
\end{equation*}
\emph{Step 3:} Apply the update step, $\delta X^*, \delta U^*$, while ensuring descent on the original nonlinear problem through the use of a merit-function and a line-search~\cite{nocedal1999numerical}.

Within that framework, Adabag et al.~\cite{adabag2024mpcgpu}, leveraged a symmetric stair preconditioner~\cite{bu2024symmetric} to solve the KKT system~\eqref{eq:KKT} through a Schur complement, preconditioned conjugate gradient, iterative linear system solve, and a parallel line search with the L1 merit function:
\begin{equation} \label{eq:merit-func}
    \mathcal{M}(X,U) = J(X,U) + \mu|c|.     
\end{equation}
This approach maximizes parallel performance on the GPU, but is customized for solving only a single problem while utilizing the entire GPU. In Section~\ref{sec:design} we develop a computational approach that leverages similar underlying algorithmic approaches but enables high-performance for batches of tens to low hundreds of solves.


\iffigs
\begin{figure*}[!t]
    \centering
    \vspace{5pt}
    \includegraphics[width=0.98\linewidth]{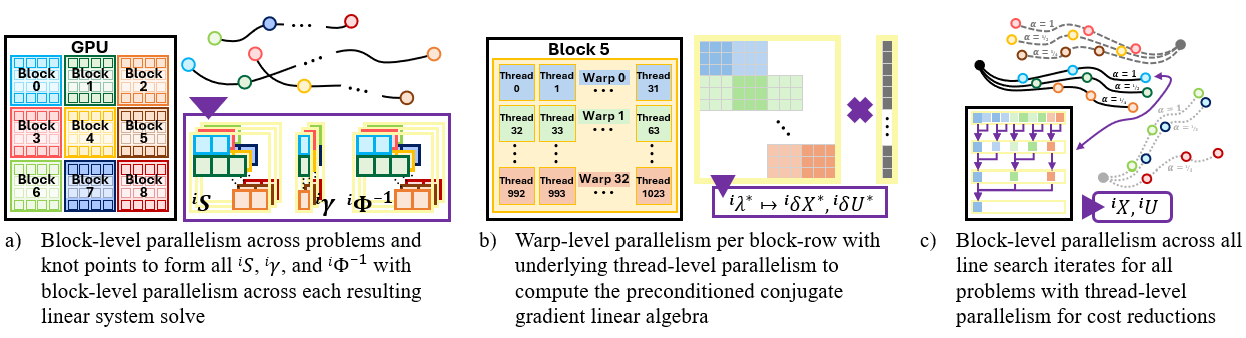}
    \vspace{-5pt}
    \caption{Overall design of our batched solver which a) forms problems in parallel across solves and timesteps, b) leverages warp-level parallelism within each block-based solve, and c) again leverages large-scale parallelism across the whole GPU for the line search and merit function calculations.}
    \label{fig:design}
    \vspace{-15pt}
\end{figure*}
\fi

\subsection{Schur Complement Iterative Methods}
\label{sec:background:schurIterative}
Iterative methods solve the problem $S\lambda^* = \gamma$ for a given $S$ and $\gamma$ by iteratively refining an estimate for $\lambda$ up to tolerance $\epsilon$. The most popular of these methods is the conjugate gradient (CG) method, which is used in the current state-of-the-art, GPU-accelerated TO solver~\cite{adabag2024mpcgpu}, and also for general, large-scale optimization problems on the GPU~\cite{naumov2011incomplete,schubiger2020gpu}.
The convergence rate of CG is directly related to the spread of the eigenvalues of $S$.
Thus, a preconditoning matrix $\Phi \approx S$ is often applied to instead solve the equivalent problem with better numerical properties: $\Phi^{-1} S\lambda^* = \Phi^{-1} \gamma$.
To do so, the preconditioned conjugate gradient (PCG) algorithm leverages matrix-vector products with $S$ and $\Phi^{-1}$, as well as vector reductions, both parallel friendly operations.

The PCG algorithm also requires the linear system $S$ to be symmetric positive definite. As such, per~\cite{adabag2024mpcgpu}, we form the \emph{Schur Complement}, $S$, and then solve~\eqref{eq:KKT} as follows:
\begin{equation} \label{eq:schur}
\begin{split}
    S &= -CG^{-1}C^T,        \hspace{20pt}      
    \gamma = c - CG^{-1}g, \\
    S\lambda^* &= \gamma,  \hspace{50.5pt}     
    \delta Z^* = -G^{-1}(g - C^T\lambda^*). \\
\end{split}
\end{equation}
By defining the variables $\theta$, $\phi$, and $\zeta$:
\begin{equation} \label{eq:trajoptSchur_1}
\begin{split}
    \theta_k &= A_k Q_k^{-1} A_k^T + B_k R_k^{-1} B_k^T + Q_{k+1}^{-1}, \\
    \phi_k &= -A_k Q_k^{-1}, \\
    \zeta_k &= -A_k Q_k^{-1} q_k - B_k R_k^{-1} r_k + Q_{k+1}^{-1} q_{k+1},
\end{split}
\end{equation}
$S$, $\gamma$, and the symmetric stair preconditioner, $\Phi^{-1}$~\cite{bu2024symmetric}, take the following forms, where $S$ and $\Phi^{-1}$ are block-tridiagonal:
\begin{equation} \label{eq:trajoptSchur_2}
\begin{split}
    S &= -\begin{bmatrix}
        Q_0^{-1} & \phi_0^T & \\
        \phi_0 & \theta_0 & \phi_1^T \\
        & \phi_1 & \theta_2 \\
        & & \ddots \\
    \end{bmatrix}_{\textstyle \raisebox{2pt}{,}} \\
    \gamma &= c - \begin{bmatrix} Q_0^{-1}q_0 & \zeta_0 & \zeta_1 & \dots & \zeta_{N-1} \end{bmatrix}^T, \\
    \Phi^{-1} &= -\begin{bmatrix}
        Q_0 & -Q_0 \phi_0^T \theta_0^{-1} &  \\
        -\theta_0^{-1} \phi_0 Q_0 & \;\; \theta_0^{-1} \;\; & -\theta_0^{-1} \phi_1^T \theta_1^{-1} \\
         & -\theta_1^{-1} \phi_1 \theta_0^{-1} & \theta_1^{-1} \\
        & & \ddots
    \end{bmatrix}_{\textstyle \raisebox{2pt}{.}}
\end{split}
\end{equation}

\section{Design and Implementation} \label{sec:design}
In this section, we describe the design of GATO as visualized in Figure~\ref{fig:design} and described in the pseudocode in Algorithm~\ref{alg:gato}. The solver architecture is optimized for GPU-parallel computations across tens to low-hundreds of trajectory optimization solves ($M$), each with tens to low-hundreds of timesteps ($N$). This paradigm, as noted in the introduction, is commonly found across robotics applications and is underserved by current solvers.

Our overall design is inspired by MPCGPU~\cite{adabag2024mpcgpu} and leverages a similar GPU-first architecture and overall 3-step design flow. However, while MPCGPU is customized for single-solve performance, we designed a new underlying solver to enable high-performance parallelism across multiple solves without sacrificing solver accuracy. We also implemented a number of additional fine-grained parallelism optimizations both across and within solves. As shown in Section~\ref{sec:results} this improves performance across all batch sizes.  

At a high level, our design leverages block-based parallelism to divide up discrete naturally parallel components of each step of our batched solve. Depending on the stage of the solver this either happens at the timestep level or problem level. Within each block we leverage warp- and thread-level intrinsics and parallelism to maximize performance and minimize overheads. Finally, by moving all of the computation onto the GPU we avoid costly I/O penalties. In the remainder of this section we detail our design.

\subsection{Batched Problem Setup and Line Search} \label{sec:design:other}
GATO is designed to maximize all possible parallelism arising from the computational structure of the underlying (batched) optimal control problems. This is most apparent in the problem setup and line search steps (shown as steps a and c in Figure~\ref{fig:design}). Here we must form $S$, $\gamma$, and $\Phi^{-1}$ per equation~\ref{eq:trajoptSchur_2} and solve a line search for the final update to $Z$, namely $Z \gets Z + \alpha^*\delta Z^*$, under a merit function, $\mathcal{M}$:
\begin{equation} \label{eq:line-search}
    \alpha^* = \text{arg}\min\limits_{\alpha_i} \mathcal{M}(Z+\alpha_i\delta Z^*) \quad \alpha \in [\sfrac{1}{\beta^0},\ldots,\sfrac{1}{\beta^\mathcal{A}}],
\end{equation}
where $\beta>1$ and $\mathcal{A}$ are positive integer values, with $\mathcal{A}$ representing the number of line search iterates.
Throughout the steps, we compute gradients and Hessians of the costs and dynamics functions across all problems and timesteps ($N*M$ total timesteps), as well as compute the merit function values to support our line search, again requiring underlying cost and dynamics calculations across all problems, timesteps, and line search iterates ($N*M*\mathcal{A}$ total timesteps). As such, we exploit block-based parallelism for each timestep to maximize both the independent nature of these computations, as well as opportunities for within-computation thread-based parallelism for the underlying small-scale linear algebra. We use the GRiD~\cite{plancher2022grid} library for efficient dynamics (gradient) computations, which follows a similar computational model.

\floatstyle{spaceruled}
\restylefloat{algorithm}
\begin{algorithm}[!t]
\begin{spacing}{1.25}
\begin{algorithmic}[1]
\caption{GATO ($X_{init}, U_{init}, N,M,\mathcal{A}, \rightarrow X^*, U^*$)}
\label{alg:gato}
\For{$b = 0 \ldots N*M$} \textbf{in parallel blocks}
    \State $S_b,\gamma_b,\Phi^{-1}_b$ via \eqref{eq:trajoptSchur_2} with \textbf{parallel threads}~(\ref{sec:design:other})
\EndFor
\For{$b = 0 \ldots M$} \textbf{in parallel blocks}
    \State $\delta Z^*$ via \eqref{eq:schur} with \textbf{parallel warps of threads}~(\ref{sec:design:pcg})
\EndFor
\For{$b = 0 \ldots N*M*\mathcal{A}$} \textbf{in parallel blocks}
    \State $\mathcal{M}_b$ via \eqref{eq:merit-func} with \textbf{parallel threads}~(\ref{sec:design:other})
\EndFor
\For{$b = 0 \ldots M$}
    \State $\alpha^*_b$ via \eqref{eq:line-search} with \textbf{parallel threads}~(\ref{sec:design:other})
\EndFor
\State \Return $X^*, U^*$
\end{algorithmic}
\end{spacing}
\end{algorithm}

Importantly, because all computations to form $S$, $\gamma$, and each timestep's merit function are block-local, only cheap \emph{intra}-block synchronizations are needed. Only a single \emph{grid-wide} synchronization is required to finalize $\Phi^{-1}$, and a block-level reduction is used to compute the merit function for each line search iterate across all batches of solves.

Throughout these computations, temporary variables are computed in fast shared memory and all final matrices and vectors are arranged densely and contiguously in global memory to maximize naturally coalesced loads and stores by the downstream PCG solver. We also reserve the device’s persisting L2 cache to reduce global memory access during PCG. Most importantly, only the current system state(s) and goal(s), as well as the final optimized state and control trajectories, incur round-trip CPU-GPU data transfer overheads.

\subsection{Batched PCG} \label{sec:design:pcg}
A key factor of GATO's overall performance is our batched linear system solver which is built around per-block PCG solves with finer-grained warp-level\footnote{We note that a ``warp'' represents 32 contiguous threads on the same GPU-core. These threads work in lock-step due to the design of NVIDIA GPU hardware. By exploiting their native implicit synchronization at the hardware level, further acceleration of software can be achieved.} parallel linear algebra. This hardware-optimized design not only improves computational throughput, but also improves memory access patterns over MPCGPU, reduces synchronizations, and increases overall hardware resource utilization both for a single solve and, most importantly, for batches of solves.

Each CUDA thread block is assigned to solve one linear system~\eqref{eq:trajoptSchur_2}. Within a block, warps distribute work over knot points, and individual threads within a warp operate on rows/columns of the per-knot state/control blocks. This mapping eliminates inter-block coordination entirely: all vector updates, matrix-vector multiplications, and local reductions are resolved inside the block, avoiding the use of intra-block synchronization, e.g., the \texttt{cooperative groups} API used in~\cite{adabag2024mpcgpu}. This design improves both per-solve performance and portability across devices and launch contexts. This is because intra-block APIs require all blocks to be co-resident on the GPU, which constrains scalability and is particularly limiting on edge systems with restricted hardware resources. 

All matrices/vectors are packed contiguously in row-major order by batch and knot points, exploiting the block tridiagonal structure of the $S$ and $\Phi^{-1}$ matrices. This yields coalesced loads/stores for warp-strided accesses and makes warp shuffle intrinsics efficient for reductions. We also pad leading dimensions to multiples of the warp size to remove bounds checks and branch divergence. This enables us to implement a \emph{warp-optimized} block tridiagonal matrix-vector multiplication routine that (i) uses shared memory tiles to stage the current and neighboring blocks, (ii) performs thread-parallel fused multiply–adds for the block-dense operations, (iii) pipelines loads to hide any memory latency (through the use of \texttt{cudaMemcpyAsync}), and (iv) avoids atomics or grid-wide barriers. The CUDA kernel’s shared-memory footprint, block dimensions, and register usage are also tuned to maintain high occupancy while preventing register spilling for typical state/control sizes seen in MPC applications. As a result, each PCG solve proceeds efficiently and fully independently. 

Finally, we partially unroll all inner loops over small, compile-time dimensions to reduce loop overhead, and compile with aggressive optimization flags (e.g., \texttt{-O3}, \texttt{-use\_fast\_math}).
As shown in Section~\ref{sec:results}, these choices result in superior performance across our target batch sizes.

\section{Results} \label{sec:results}
In this section, we present a two-part evaluation of GATO. We first test our solver on a number of software benchmarks aimed to evaluate our approach against relevant baselines and explore the scalability of our design. We then demonstrate the usefulness of our solver through case studies of batched trajectory optimization for MPC applications. Our case studies are first demonstrated via simulation ablations. The final case study is also deployed onto a physical KUKA iiwa LBR14 manipulator. The source code accompanying this evaluation is released open source alongside our solver.


\subsection{Methodology} \label{sec:results:method}
Results were collected on a high-performance workstation with a $5.73$GHz 24-core AMD Ryzen 9 7900X i9-12900K and a $2.2$GHz NVIDIA GeForce RTX 4090 GPU running Ubuntu 22.04 and CUDA 12.6. Code was compiled with \texttt{g++11.4}, and time was measured using high-precision \texttt{timeit} package around \texttt{Python} wrappers for all CPU and GPU functions to provide realistic timing analysis for future users of our open source software. 

Throughout our experiments, we compare our solver to ablations of itself, the state-of-the-art CPU QP solver OSQP~\cite{stellato2020osqp} using the Pinocchio~\cite{carpentier2019pinocchio} dynamics library, and the state-of-the-art GPU solver MPCGPU~\cite{adabag2024mpcgpu} using the GRiD dynamics library~\cite{plancher2022grid}. We note that we also leverage the GRiD library in GATO as mentioned in Section~\ref{sec:design}. All hyperparameter values can be found in our open source code. All solvers used the same cost functions, and solver-specific hyperparameter values were independently optimized.

We exclude \texttt{Jax}-based GPU solvers (e.g.,~\cite{tracy2024differentiability,amatucci2025primal}) from our evaluations as both from their reported results in papers, and from our own evaluations on our computational hardware, they take tens of milliseconds to solve small batches of trajectory optimization problem: often an \emph{order of magnitude slower} than our baselines. Similarly, OSQP's GPU backend is known to not be performant at our target problem sizes~\cite{adabag2024mpcgpu,schubiger2020gpu} and is as such similarly excluded.


\subsection{Scalability Benchmarks} \label{sec:results:benchmark}

\iffigs
\begin{figure*}[!t]
    \centering
    \vspace{5pt}
    \includegraphics[width=0.52\linewidth]{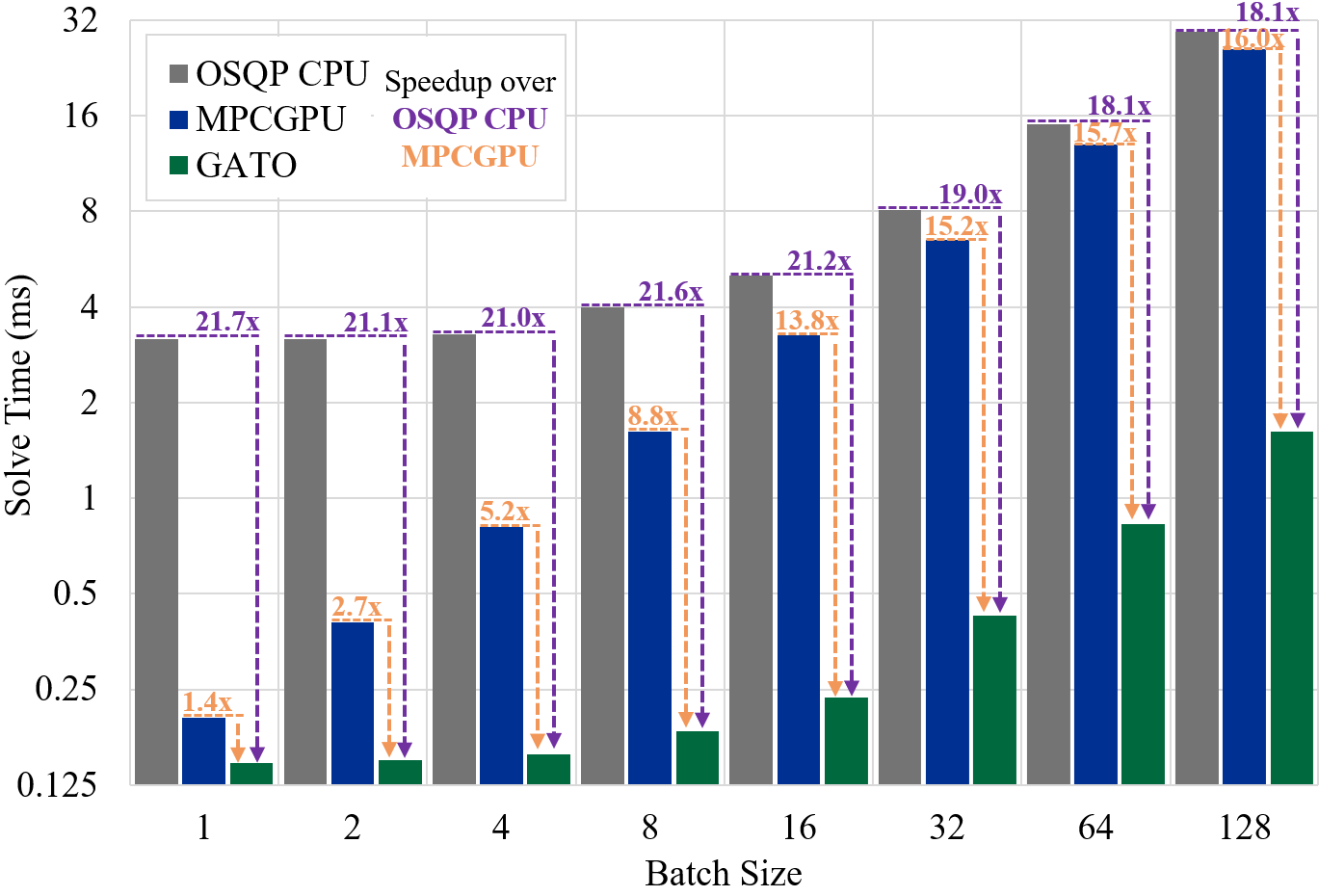} 
    \hfill
    \includegraphics[width=0.44\linewidth]{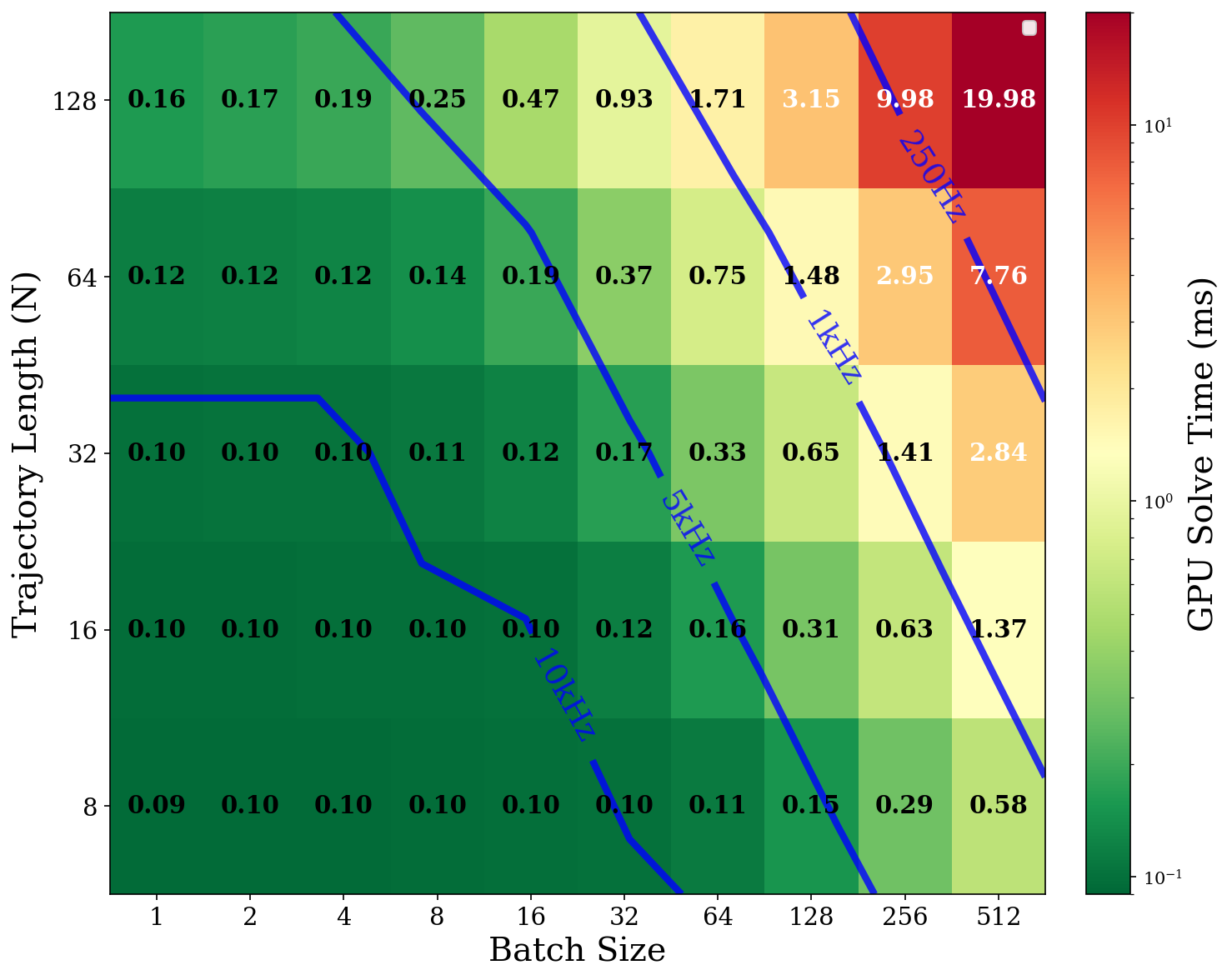}
    \caption{(Left) Solve times for 6-DoF manipulator motions while varying the batch size ($M$) and underlying solver. $N=64$ for all solves. GATO shows far improved scalability as compared to state-of-the-art CPU and GPU solutions. (Right) A heat map of solve times while varying both batch size ($M$) and time horizon ($N$). GATO is able to reach kHz control rates for real-time iterations of large batches (512) of short horizon ($N=8$) trajectories, as well as smaller batches (32) of longer horizon trajectories ($N=128$), showing the flexibility of the design.}
    \label{fig:benchmark}
    \vspace{-10pt}
\end{figure*}
\fi

We begin with a scalability study on a 6-DoF Neuromeka Indy7 manipulator executing a figure-8 tracking task. At each control step, we solve a batch of $M$ trajectory-optimization problems with a fixed horizon of $N{=}64$, $h{=}0.01\,\text{s}$, warm-started with the previous control step’s solution. Figure~\ref{fig:benchmark} (left) summarizes these results for $M = [1,2,4,\ldots,128]$ comparing GATO against the aforementioned OSQP CPU baseline and MPCGPU GPU baseline. OSQP never matches the single-problem latency of either GPU method and, while its runtime scales reasonably with problem size, it is consistently the slowest. On the other hand, while MPCGPU is just 1.4$\times$  slower than GATO for a single instance, since it is engineered to occupy the full GPU per solve, MPCGPU's latency grows nearly linearly with batch size, eventually falling behind GATO by a factor of 16$\times$ for batch size $M=128$.
Overall, GATO achieves both lower single-solve latency and stronger scaling than baselines across our target range of batch-sizes. This yields an overall $18-21\times$ speedup over our CPU baseline and $1.4-16\times$ over our GPU baseline.

Figure~\ref{fig:benchmark} (right), shows a heat map of solve times for GATO while varying both batch size ($M$) and time horizon ($N$) for the same tracking problem. GATO is able to reach kHz control rates for real-time iterations of large batches ($M=512$) of short horizon ($N=8$) trajectories, as well as smaller batches ($M=32$) of longer horizon trajectories ($N=128$), showing the flexibility of its design. We also find that this scalability is mostly related to the total number of knot points in the overall problem ($N * M$). For example, all points under 512 total points, e.g., $(N,M) = (64,8), (8,64), (16,32)$, can run a real-time iteration in about 100$\mu$s (10kHz control rate) indicating that our solver efficiently utilizes GPU resources up to hardware limits, and then scales linearly for subsequent increases in problem size. We note that this surpasses the 512 point 1kHz-scaling shown in MPCGPU~\cite{adabag2024mpcgpu} and similar 1kHz max-scaling in other CPU-based state-of-the-art results~\cite{kleff2021high}. 

In the next sections, we present three case studies that demonstrate the practical value of GATO's ability to solve batches of tens–to-hundreds of TO problems in real-time. 


\subsection{Case Study 1: Online Hyperparameter Optimization}
\label{sec:results:case1}

\begin{figure}[!t]
    \centering
    \includegraphics[width=1\linewidth]{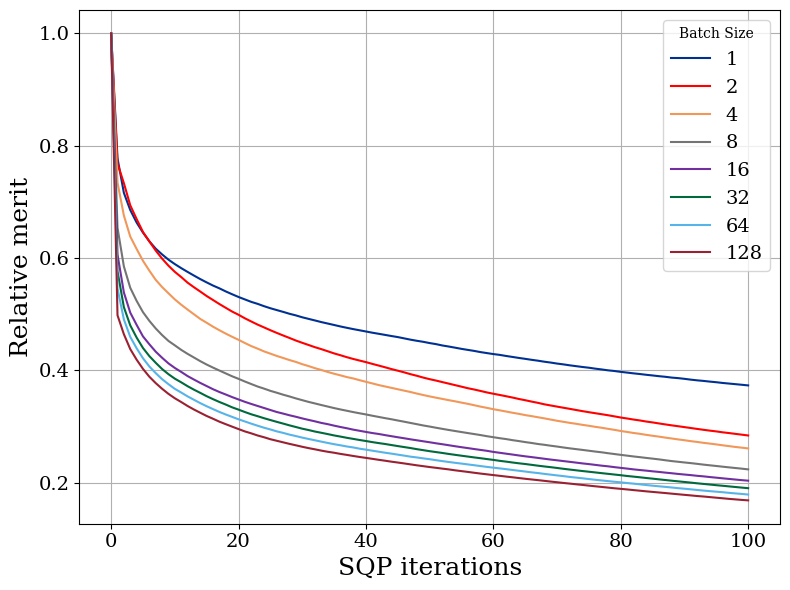}
    \caption{Average (normalized) merit function value across SQP iterations over 100 runs each with 81 different random values for the cost function parameters $Q$ and $R$ in~\eqref{eq:trajoptSchur_1}. For all solves $N=64$, $h=0.05$, $\rho$ ranges from $10^{-8}$ to $10^{1}$.}
    \label{fig:caseStudyRho}
    \vspace{-15pt}
\end{figure}

Our first case study addresses hyperparameter selection in MPC, traditionally a time-consuming and sensitive process. We consider motion planning for the 7-DoF KUKA iiwa LBR14 with horizon $N=64$ and timestep $h=0.05\,\text{s}$, running the solver for 100 SQP iterations from zero-initialized states and controls on 100 randomly sampled points within the robot's workspace. 

Our batched solver is used to sample over $\rho$, a damping parameter often added to the diagonal of $Q_k$ in~\eqref{eq:trajoptSchur_1} in deployed trajectory optimization solvers to improve numerical stability. We initialize the single-solve baseline with $\rho=10^{-1}$. For batch size $M$, we initialize $\rho$ by log‑spacing values between $10^{-8}$ and $10^{1}$. In both cases, $\rho$ is adjusted after each SQP iteration based on the status of the line search, similar to the scheme in \cite{transtrum2012}.

Figure~\ref{fig:caseStudyRho} shows that larger batches consistently reduce merit function values faster (indicating faster convergence to an optimal solution). Batches larger than 16 outperform the minimum merit achieved by the single-solve baseline after only 20 iterations, and $M=32$ through $128$ achieve nearly half the initial merit after only a single SQP iteration. However, we observe that these gains begin to saturate beyond $M \approx 32$ due to the limited range in $\rho$, reconfirming the importance of batch sizes of tens to low-hundreds. Overall, this improved convergence rate would enable a deployed solver to achieve comparable optimality at higher control rates, or increased optimality at a set nominal control rate.

\iffigs
\begin{figure*}[!t]
    \centering
    \vspace{4pt}
    \includegraphics[width=0.47\linewidth]{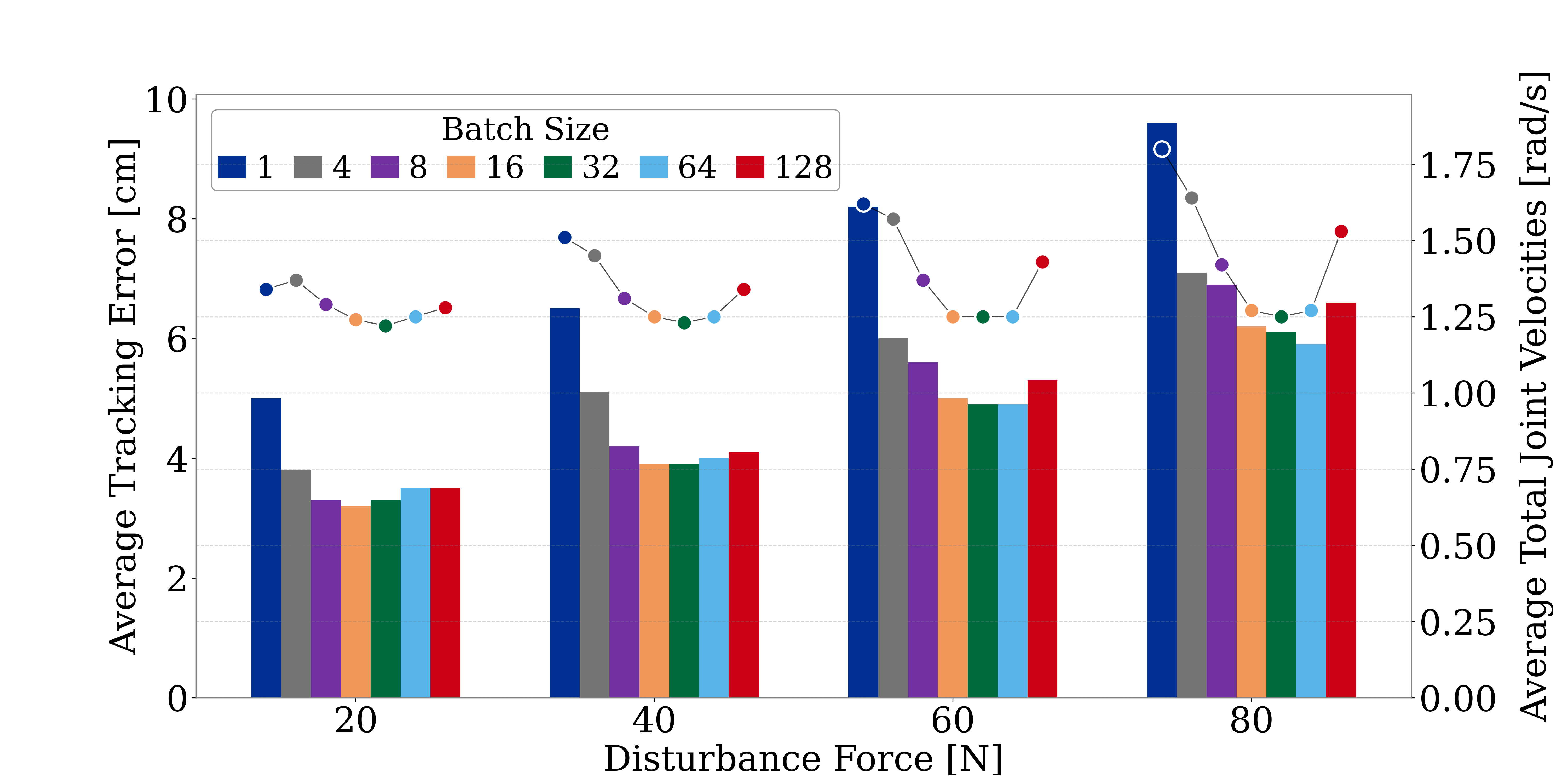} 
    \hfill
    \includegraphics[width=0.52\linewidth]{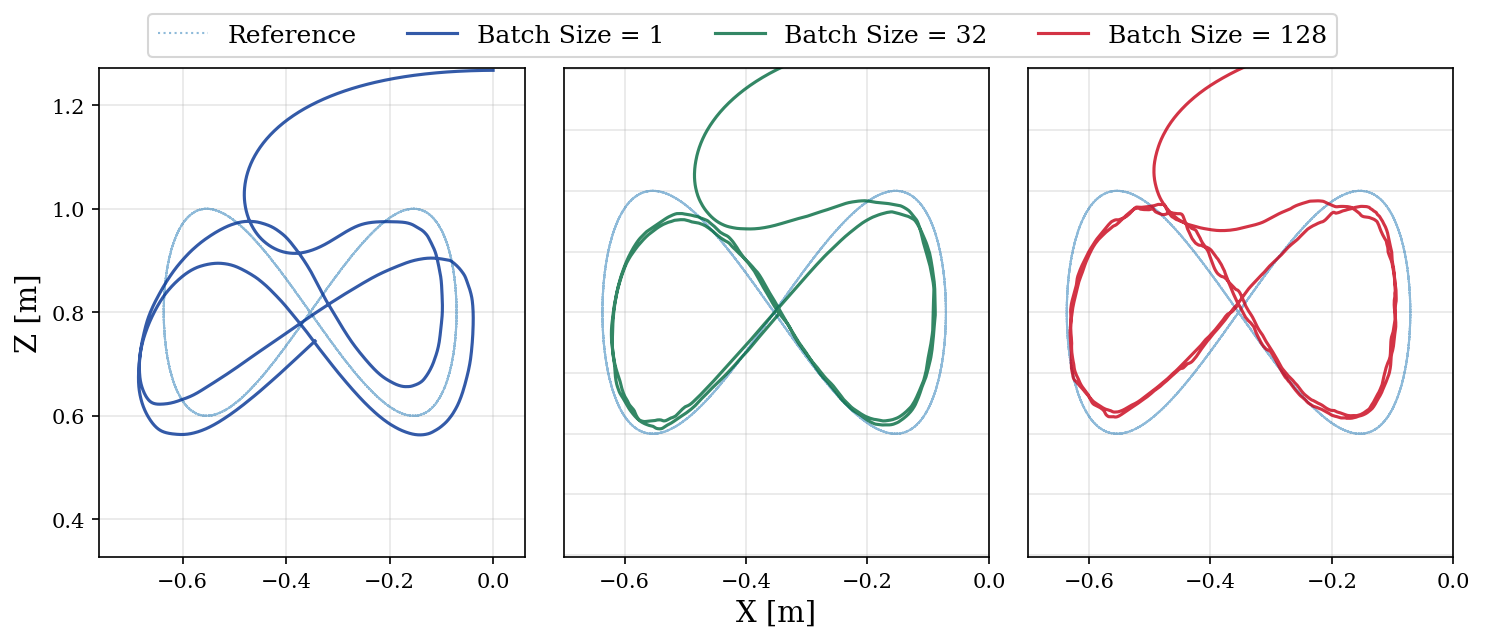}
    \caption{Figure-8 tracking task, with an external disturbance applied at the end effector. (Left) Bar chart shows tracking error, scatter plot shows average total joint velocities. Increasing GATO's batch size enables increased disturbance reject, lowering tracking error and joint velocities until the increased latency from a larger batch size outweighs the optimality gains. (Right) End-effector trajectories realized during this experiment when 50N of external force is applied at the end effector, again showing that modest batch sizes lead to the best performance.}
    \label{fig:caseStudyFixed}
    \vspace{-14pt}
\end{figure*}
\fi

\subsection{Case Study 2: Fixed Disturbance Rejection}
\label{sec:results:case2}
Our second case study explores disturbance rejection, a common problem in robotic control tasks.
Here, a 6-DoF manipulator tracks a figure-8 end-effector trajectory like in \ref{sec:results:benchmark}, but now faces an unmodeled constant external force applied at the end effector in the $-Z$ direction. 

Batched TO enables an ``online hypothesize-and-test'' strategy: evaluate multiple candidate disturbance models in parallel and apply the control from the most consistent one. At each control step, we solve a batch of $M$ TO problems differing only in the assumed external force, $f_j \;\text{for}\; j\in[0,M)$. Candidate forces are generated by sampling directions uniformly on a sphere and adding them to the current estimated disturbance, exploring both direction and magnitude around the prior hypothesis. After solving this batch of problems, we use the optimized trajectory whose dynamics model best matches the measured evolution of the robot's state after one control step. We then update our disturbance estimate for our next solve by re-centering it around the selected $f_j$.

As shown in Figure~\ref{fig:caseStudyFixed}, this simple sampling approach proves effective, consistent with batched roll-outs as noted in~\cite{howell2022predictive} and batched contact estimates as noted in~\cite{suh2022bundled}. In particular, Figure~\ref{fig:caseStudyFixed} (left) shows that tracking error and joint velocities decrease with increasing $M$ until reaching a sweet spot at around $M = 32$. Beyond this point, increased solve times offset the benefit of finer hypothesis granularity and increase closed-loop error. Figure~\ref{fig:caseStudyFixed} (right) illustrates end-effector trajectories for a representative 50 N disturbance, where $M=32$ tracks the figure-8 substantially better than a single-solve baseline, while very large batches, e.g., the $M=128$ shown, lose effectiveness due to higher latency.


\subsection{Case Study 3: Planning Under Uncertainty}
\label{sec:results:case3}

\begin{figure} [!t]
    \centering
    \includegraphics[width=1\linewidth]{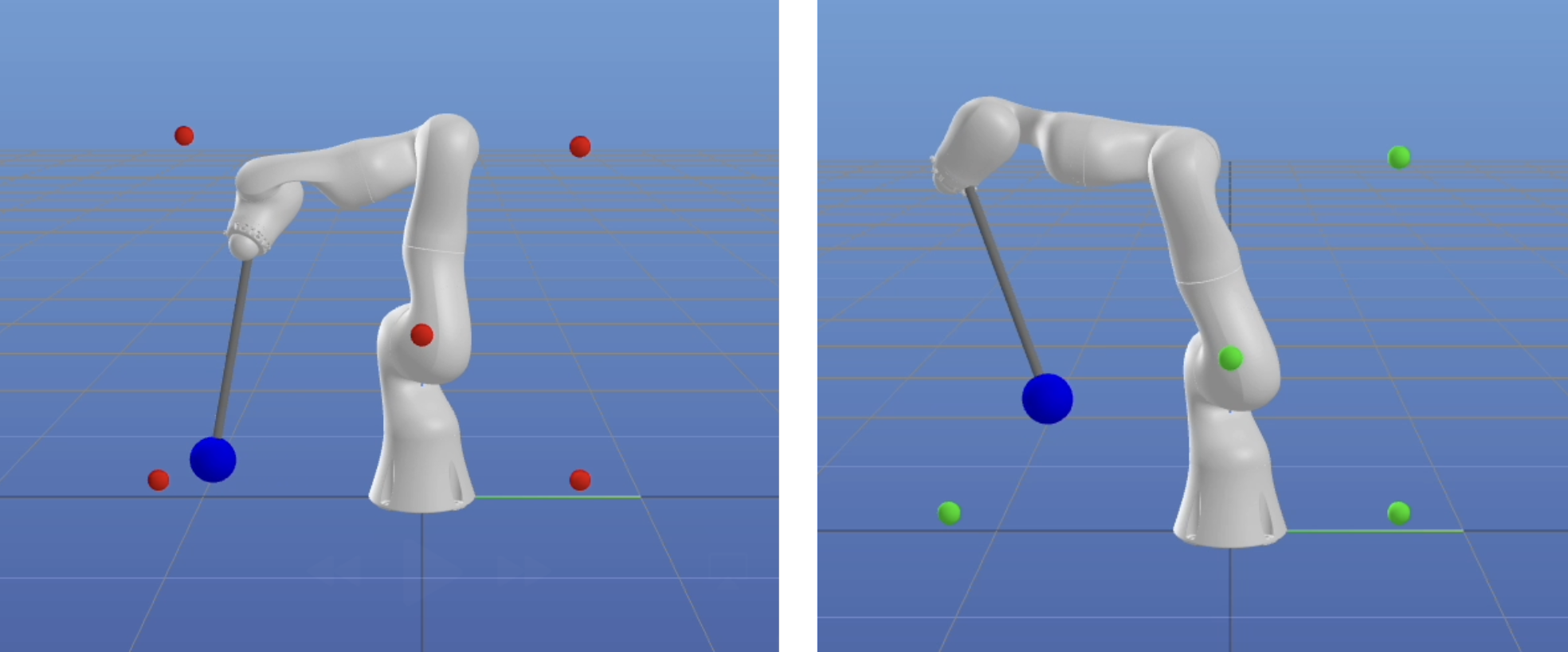}
    \caption{Simulation visualization at the last timestep of the pick-and-place task in Section~\ref{sec:results:case3}, with a 15kg pendulum attached to the last joint. Batch size $M=1$ on the left, and $M=32$ on right.}
    \label{fig:case study 3 sim visualization} 
    \vspace{-15pt}
\end{figure}

In our final case study, we consider a 7-DoF KUKA iiwa LBR14 executing a multi-point pick–and–place task with an \emph{unmodeled} suspended load attached to the end effector (see Figure~\ref{fig:case study 3 sim visualization}).
The swinging payload induces time-varying, direction-dependent forces that degrade controller performance. To account for this, at each control step, and as done in Section~\ref{sec:results:case2}, GATO warm-starts from the previous solution and solves a batch of trajectory-optimization problems in parallel, each conditioned on a different disturbance hypothesis. Controls are then selected or blended according to consistency with the observed motion, and the hypothesis set is re-centered for the next step. 
Throughout, the task enforces tight accuracy requirements with success requiring the end effector to reach within 5 cm of each goal in under 5 seconds. We also require the sum of joint-velocities to be under 1.0 rad/s. If time is exceeded, the target is considered a failure and we move onto the next target. This experiment demonstrates GATO's robustness and shows why our target batch sizes are practical: they offer sufficient disturbance coverage without sacrificing real-time performance. 

\subsubsection{Simulation Studies}
\label{sec:results:case3:software}

In simulation the solver uses a horizon of $N{=}16$ with a timestep $h=0.01\,\text{s}$, and is limited to 5 SQP iterations with a PCG tolerance of $10^{-6}$. We simulate the plant at $1\,\text{kHz}$ (RK4 with $h{=}0.001\,\text{s}$). We use a constant 15kg mass and run 100 scenarios varying pendulum length $\ell\in[0.3,0.7]$\,m, initial angle $\|\theta\|\in[0,0.6]$\,rad, and damping constant $b\in[0.1,0.6]$\,Nms/rad.

\begin{figure}[!t]
    \centering
    \vspace{5pt}
    \includegraphics[width=1\linewidth]{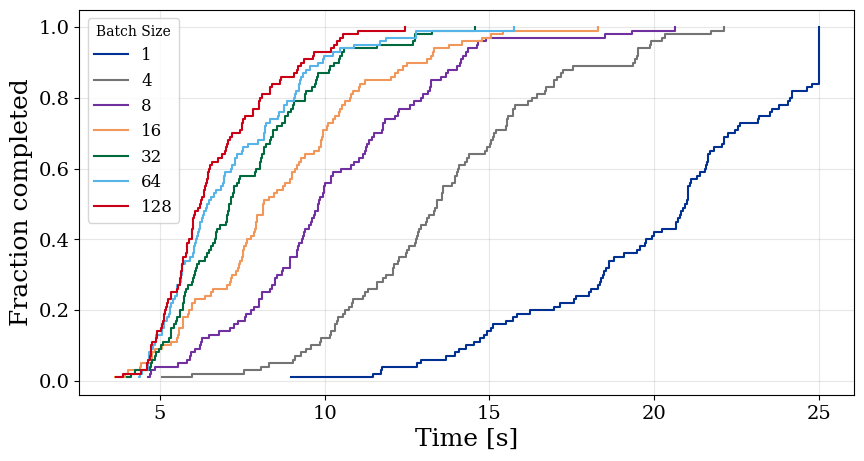}
    \vspace{-10pt}
    \caption{Cumulative density function of the solve times for a trajectory length $N{=}16$ for GATO across different batch sizes and varied pendulum configurations. For each batch size, the solver accounts for 100 disturbance scenarios. We see how larger batch sizes enable more accurate unmodeled disturbance rejection.}
    \label{fig:Case_study_3CFD}
    \vspace{-5pt}
\end{figure}

Table~\ref{tab:caseStudy3} summarizes the solver's performance and Figure~\ref{fig:Case_study_3CFD} shows the distribution of solve times for GATO across different batch sizes. We can see that the success rate dramatically increases and the task-completion time falls significantly as $M$ grows from 1 to 8. Following that, performance continues to increase albeit at a slower rate. Ultimately, at our largest batch size of $M=128$ (shown in red), we are able to not only achieve a 99.2\% success rate but also solve almost all problems faster than any other solver, with $M = 64$ and $32$ (shown in light blue and green) not far behind. Figure~\ref{fig:case study 3 sim visualization} provides a visualization of the simulated experiments, with the red and green spheres denoting unreached and reached targets respectively for $M=1$ (left) and $M=32$ (right).

\begin{table}[!t]
  \centering
  \caption{Episode performance vs batch size, collected by varying pendulum configuration (length, angle, damping).}
  \begin{tabular}{|c|c|c|}
    \hline
    Batch size & Success rate [\%]  & Mean time [s]\\
    \hline
    1 & 33.0 & 20.1 \\
    \hline
    4 & 78.2 & 13.7 \\
    \hline
    8 & 91.8 & 10.2 \\
    \hline
    16 & 95.0 & 8.7 \\
    \hline
    32 & 96.4 & 7.5 \\
    \hline
    64 & 97.6 & 7.1 \\
    \hline
    128 & 99.2 & 6.7 \\
    \hline
  \end{tabular}
  \label{tab:caseStudy3}
  \vspace{-10pt}
\end{table}

\subsubsection{Hardware Deployment}
\label{sec:results:case3:hardware}

Finally, we run two variants of the simulation experiments from \ref{sec:results:case3} on a physical KUKA iiwa LBR14 robot: (a) five-target goal reaching with no load at 100Hz, and (b) three-target goal reaching with an unmodeled 4kg load at 1000Hz. Both used horizon $N{=}32$, timestep $h=0.02\,\text{s}$, one SQP iteration, and PCG tolerance of $10^{-6}$. Our goal is to show real-world effectiveness of our GPU-accelerated approach, handling not only unmodeled forces but also control loop delays, system identification errors, and noisy sensor measurements. We compare results from a single solve against a batch size of $M=32$.

\begin{figure}[!t]
    \centering
    \vspace{5pt}
    \includegraphics[width=1\linewidth]{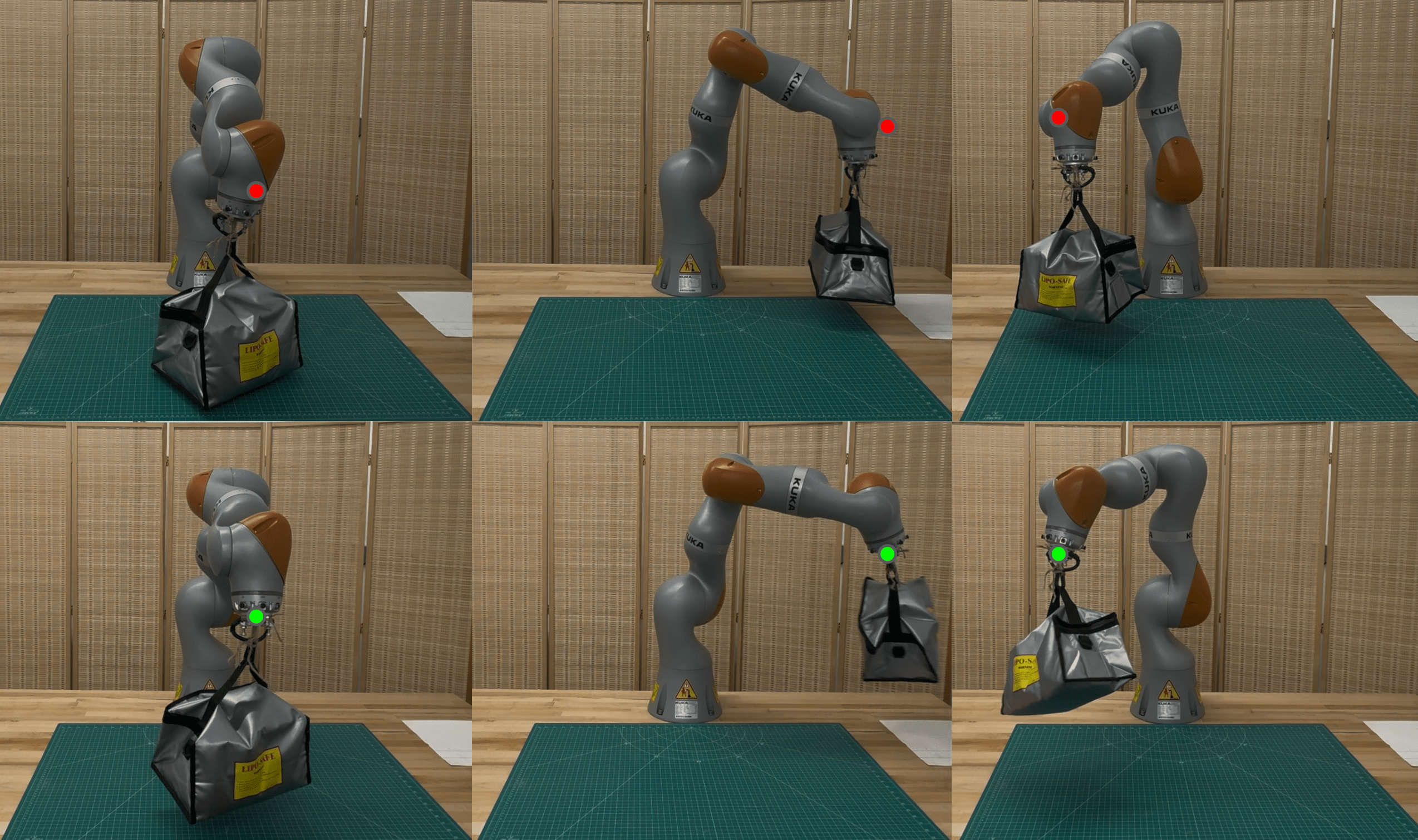}
    \caption{Hardware experiment (b) showing the solver successfully ($M=32$) and unsuccessfully ($M=1$) account for the time-varying unmodeled disturbance and reach the targets.}
    \label{fig:hardware}
    \vspace{-5pt}
\end{figure}

As shown in Figure~\ref{fig:hardware}, Table~\ref{tab:hardware-task-metrics}, and our supplementary videos, the batched solver outperforms single solves, reaching targets in less time and successfully rejecting model errors, sensor noise, and the time-varying external disturbance.

\section{Conclusion and Future Work} \label{sec:conclusion}
In this work, we introduce GATO, an open source, GPU-accelerated, batched TO solver that is co-designed across algorithm, software, and computational hardware to deliver real-time throughput for batches of tens to low-hundreds of solves. GATO achieves its performance through co-designed parallelism at the block-, warp-, and thread-level, taking full advantage of the GPU computational model. Our experiments demonstrate not only superior performance, providing speedups of as much as $18–21\times$ over CPU and $1.4–16\times$ over GPU baselines as batch size increases, but also that such moderate batch sizes are useful for deployed applications, improving convergence, and rejecting disturbances both in simulation and on a physical manipulator.

\begin{table}[!t]
    \centering
    \caption{Performance metrics for the hardware pick-and-place tasks showing the improved performance resulting from the use of our batched solver.}
    \begin{tabular}{|c|c|c|c|c|c|}
        \hline
        Experiment & \multicolumn{2}{|c|}{\textbf{\textit{(a)}}} & \multicolumn{2}{|c|}{\textbf{\textit{(b)}}} \\
        \hline
        Batch Size & 1 & 32 & 1 & 32 \\
        \hline
        Completed & 5/5 & 5/5 & 0/3 & 3/3 \\ 
        \hline
        Time [s] & 16.93 & 9.49 & 24.00 & 6.71 \\
        \hline
    \end{tabular}
    \label{tab:hardware-task-metrics}
\end{table}

There are many promising directions for future work, including: integration with actor-critic reinforcement learning to guide agent exploration~\cite{alboni2024cacto},  use of branch-and-bound-based methods for contact-implicit trajectory optimization~\cite{marcucci2020warm}, and evaluation of our approach on mobile robots at the edge using low-power GPU platforms such as the NVIDIA Jetson~\cite{ditty2022nvidia}.

\section{Acknowledgments}
We thank Ludovic Righetti and the Machines in Motion Laboratory for their support with hardware demonstrations.

\bibliographystyle{styles/IEEEtran_new}
\bibliography{styles/IEEEabrv,a2r}

\end{document}